\documentclass{article}

\usepackage{amssymb}
\usepackage{graphicx}
\usepackage{subfig}
\usepackage{url}
\usepackage{authblk}

\providecommand{\keywords}[1]
{
  \small	
  \textbf{\textit{Keywords---}} #1
}

\title{Coast Sargassum Level Estimation from Smartphone Pictures}

\author[1]{Abril Valeria Uriarte-Arcia}
\author[1,2]{Juan Irving Vasquez-Gomez \footnote{Corresponding author: jvasquezg@ipn.mx}}
\author[1]{Hind Taud} 
\author[1]{Andres Garcia-Floriano}
\author[1]{Elías Ventura-Molina}

\affil[1]{Instituto Politécnico Nacional (IPN), Centro de Innovación y Desarrollo Tecnológico en Cómputo, Av. Juan de Dios Bátiz S/N, Gustavo A. Madero, 07738, Ciudad de México, México}
\affil[2]{Consejo Nacional de Ciencia y Tecnología (CONACYT), Av. Insurgentes Sur 1582, Benito Juárez, 03940, Ciudad de México, México}

\date{}

\begin{document}
\maketitle

\begin{abstract}
Since 2011, significant and atypical arrival of two species of surface dwelling algae, \textit{Sargassum natans} and \textit{Sargassum Fluitans}, have been detected in the Mexican Caribbean. This massive accumulation of algae has had a great environmental and economic impact. Therefore, for the government, ecologists, and local businesses, it is important to keep track of the amount of sargassum that arrives on the Caribbean coast. High-resolution satellite imagery is expensive or may be time delayed. Therefore, we propose to estimate the amount of sargassum based on ground-level smartphone photographs. From the computer vision perspective, the problem is quite difficult since no information about the 3D world is provided, in consequence, we have to model it as a classification problem, where a set of five labels define the amount. For this purpose, we have built a dataset with more than one thousand examples from public forums such as Facebook or Instagram and we have tested several state-of-the-art convolutional networks. As a result, the VGG network trained under fine-tuning showed the best performance. Even though the reached accuracy could be improved with more examples, the current prediction distribution is narrow, so the predictions are adequate for keeping a record and taking quick ecological actions.
\end{abstract}


\section{Introduction}

Since 2011, significant and atypical arrival of two species of surface dwelling algae, \textit{Sargassum natans} and \textit{Sargassum Fluitans}, have been detected in the Mexican Caribbean Sea \cite{Munoz2019}. This Great Atlantic Sargasso Belt, named by \cite{Wang2019}, estimated maximum coverage was 6,000 $km^2$ in 2018 containing approximately 20 million metric tons of Sargassum biomass \cite{Wang2019}. According to \cite{Chavez2020}, the most affected area in Mexico is the coastline between Tulum and Playa del Carmen including the east coast of Cozumel Island. This massive accumulation of algae has had a great environmental and economic impact, especially for the tourism sector, which is an activity of great importance for the inhabitants of the Mexican Caribbean. In addition to the economic impact on tourism, it can also have a strong impact on the ecosystem. Although these floating masses of Sargassum are generally considered beneficial to the biodiversity of the high seas, the effect of unusually large masses remains to be seen. Apart from the unpleasant smell and negative visual impact on the beach, possible ecosystem impacts are: beach erosion due to removal efforts; decomposing Sargassum can induce temperatures that are lethal for turtle embryos; newly hatched sea turtles can also be affected on their way to the sea; nearshore benthic fauna (including coral colonies) and seagrasses can be affected by the change in light condition, oxygen level, temperature, and water pH \cite{Chavez2020}.

Most of the works related to algae detection in the ocean have been carried out using remote sensing since it has proven to be a very useful technique for Earth observation. Several research efforts relate to monitoring and detecting Sargassum have been developed in recent years using data from satellites such as Landsat-8 or Sentinel-2 \cite{Arellano-Verdejo2019,Balado2021,Cuevas2018,Marechal2017,Sun2021,Wang2017}.

Although the fact that remote sensing is a powerful tool, it still has some disadvantages for treating certain problems, being the main disadvantages the spatial and temporal resolution. For instance, Sentinel-2 has a spatial resolution of 10, 20 and 60 m. \cite{Sentinel2}, which indicates the surface area measured on the ground represented by an individual pixel. This resolution allows to cover extensive areas in one image, but for problems like Sargassum detection in coastal areas a higher spatial resolution, despite covering smaller areas, could be more appropriate. The temporal resolution is an even more serious problem, as an example, Sentinel-2 has temporal resolutions of 10 and 5 days, which means that it will take 10 or 5 days before the satellite revisit a particular point on the Earth's surface. Sargassum observation in beach areas requires a more frequent observation of the phenomenon. Another important factor is the meteorological condition in the area of interest, under cloudy conditions during the raining/hurricane season, the information provided by the satellite images would be limited. 

Given the previously mentioned conditions, ground-based camera systems can be more adequate to address the problem at hand. This approach has been little explored \cite{Arellano-Verdejo2020,Valentini2020}, but the few studies carried out have shown promising results. Another aspect that can be exploited is the rapid development of mobile technologies, which has led to the high availability of high-resolution images taken from cell phones.

In this work, we propose an approach for a sargassum amount estimation. We model the problem as a classification task where a set of five labels define the amount in each image. The classification is solved with a supervised learning approach where a convolutional neural network model is trained with several already labeled examples to predict the class on unseen examples \cite{Christin2019}. In that sense, we have built a dataset with more than 1000 images from public social network forums such as Facebook and Instagram. Also, in our experiments, we have compared the performance of four state-of-the-art convolutional neural networks. As a result, our system can provide accurate predictions that can help with the sargassum monitoring. 

The rest of the paper is organized as follows. Section \ref{sec:related work} provides a review of methods that have been proposed to deal with the monitoring of sargassum. Section \ref{sec:mat and methods} describes the neural networks that are being used in this study. Then, in section \ref{sec:sargassum classification}, we detail the proposed method composed by a dataset and a classification methodology. In section \ref{sec:experiments}, we describe the experiments that were carry out for validating our proposal. Finally, in section \ref{sec:conclusions}, we provide the conclusions of our study.

\section{Related Work}
\label{sec:related work}

Giving the importance of monitoring and assess the impact of sargassum in coastal ecosystems, automated detection of sargassum is a research topic that has been investigated for several years.

The success of the application of pattern classifiers often depends on the amount of data with which it has been trained and validated. In \cite{Alvarez-Carranza2019}, the process followed for the creation of a geospatial dataset using MODIS (Moderate-Resolution Imaging Spectroradiometer) sensor data is described. The created dataset contains information of the coastline of the state of Quintana Roo from 2014 to 2018, and according to the authors is suitable for the analysis of physical and biological variables in the Caribbean Sea, highlighting its application in the monitoring of sargassum.

Remote sensing has been an important tool for this task. Authors have proposed methods for the automatic detection based on satellite images. In \cite{Chen2019}, the authors proposed a detection method based on spectral and texture features. In that work, four GLCM (Gray Level Co-occurrence Matrix) measurements and sampling spectrum from typical pixels of high-resolution satellite images were calculated. 10-dimensional patterns were created from the information of the GLCM, the sampling spectrums, and first principal component. These patterns were used to train an Support Vector Machine (SVM) with a Radial Basis Function (RBF) kernel.

Some authors have proposed the use of indexes that can be directly calculated from spectral images. In the work by \cite{Sutton2019}, images obtained with MSI/Sentinel-2, OLCI/ Sentinel-3 and MODIS/Aqua sensors are used to calculate a new index called NFAI (Normalized Floating Algae Index) which works at the highest pixel resolution namely 250m for MODIS, 300 meters for OLCI, and 20 m for MSI. An updated version of this method was proposed by \cite{Wang2020}, using FAI (Floating Algae Index) images obtained from the Multispectral Instruments (MSI) of the Sentinel-2, a series of filters were applied to denoise the images and eliminate possible occlusions due to the presence of clouds. The authors report that their method has a high performance (F-score of 86\%) when filtering the images. It should be noted that because multispectral images offer a large number of features, it is possible to propose several indices for locating sargassum in images. In \cite{Gao2018}, the authors proposed the Floating Vegetation Index (FVI) which is based on values between the 1.0 $\mu$m  and 1.24 $\mu$m bands. Test data were collected by NASA JPL AVIRIS instruments in the Gulf of Mexico area and in the south of the City of San Francisco. This index can only be calculated with recent generation sensors and has not yet been tested with a large amount of data. Four vegetation indexes and one floating algae index were proposed in \cite{Cuevas2018}. The indexes were applied to Landsat 8 imagery, and then a Random Forest algorithm was used for classification.

More recently, the use of deep neural networks has proved to be an efficient tool to address different ecological problems \cite{Christin2019}, such as identify species, classify animal behaviors and estimating biodiversity. In \cite{Arellano-Verdejo2019}, a model called ERISNet, which used convolutional and recurrent neural networks architectures, is proposed to detect this macroalgae using Aqua-MODIS imagery from the coastal zone of Quintana Roo. In \cite{Arellano-Verdejo2020}, a different approach was tested, using images from smartphones and highlighting the contribution of crowdsourcing. In this work a pre-trained AlexNet neural network (for the ImageNet dataset) was used to classify images of several regions in the state of Quintana Roo. A semantic segmentation of macroalgae is proposed in \cite{Balado2021}, three convolutional neural networks (MobileNetV2, ResNet18, Xception) were compared, being ResNet18 the one with best results. In \cite{Valentini2020}, another approach using images from smartphones is proposed. The semantic segmentation of morphological/natural regions from coastal images is performed with a combination of a convolutional neural network and sticky-edge adhesive superpixels algorithm.

\section{Materials and Methods}
\label{sec:mat and methods}

\subsection{Deep Convolutional Neural Networks}

Convolutional neural networks (CNN) are a special group of neural networks which use convolution filters for learning and extracting features from the network inputs \cite{lecun2015deep}. Once trained, each filter identifies a particular feature from the input. The filters are reused across the image reducing the number of parameters (weights) of the network. Concatenating filter layers create the capacity of detecting non-linear features, in consequence, deep CNN  can find and classify complex concepts. Next, we will review some of the state-of-the-art CNNs that will be used for experimentation.

\subsubsection{AlexNet}


The Alexnet model, proposed by Alex Krizhevsky in collaboration with Ilya Sutskever and Geoffrey Hinton \cite{Krizhevsky2012} consists of five convolutional layers and three fully-connected layers; among the innovations introduced in this model are: the use of the ReLU function instead of the tanh function; the facility to run the training using multiple GPUs, which reduces the training time; pooling overlapping is possible. Since AlexNet handles up to 60 million parameters, it can be prone to overfitting, so Data Augmentation and Dropout strategies were introduced. In 2012 AlexNet obtained a top-1 error of 0.375 and a top-5 error of 0.17, this model is able to recognize objects that are not centered and even this model won the ImageNet competition with a top-5 error of 0.153.

\subsubsection{Google Net}

This name is given to an architecture that is a variant of the Inception Network, which was presented at the ILSVRC14 (ImageNet Large-Scale Visual Recognition Challenge 2014). Among the tasks that can be developed with this model are image classification, object detection, object recognition, object classification, facial recognition, among others such as adversarial training or model compression. 

This model is composed of 27 layers: 22 convolutional layers, 5 pooling layers; these are grouped in 9 inception modules. The inception modules consider or contemplate 1x1, 3x3, 5x5 and Max Pooling convolution operations, which are concatenated and form an output. The essential idea of this architecture is to provide an efficient computation in comparison with similar or previous networks, for this the input image is sized at 224x224, which implies a size reduction, but preserving the important spatial information. In this model, a droput layer is used before the fully connected layer, this in order to prevent network overfitting, finally the fully connected layer is formed by 1000 neurons that correspond to the 1000 ImageNet classes and the final activation function is the softmax\cite{Szegedy2015}. 

\subsubsection{VGG}

The VGG architecture was introduced by \cite{SimonyanZ15} which has been very successful when applied on ImageNet dataset \cite{ILSVRC15} composed by 1,000 classes and more than one million of examples. Their main contribution consists of increasing the Convolutional Neural Network architecture depth using (3×3) convolution filters. Significant improvement on the prior-art architecture has been achieved by increasing the depth to 16 or 19 weight layers. 
Typically, this architecture is divided into five blocks of convolutions alternated by “max pooling” layers and ends with a classification block made up of densely connected layers. Several VGG models have been emerged, varying in their number of convolution layers. For example, the VGG16 contains the following convolution blocks: 2×conv3-64, 2×conv3-128, 3×conv3-256, 3×conv3-512 and  3×conv3-512. In VGG19, the third, forth and fifth block are increased by a convolution layer which comprises 2×conv3-64, 2×conv3-128, 4×conv3-256, 4×conv3-512 and 4×conv3-512. Both are ended by 3 densely connected layers.  

\subsubsection{ResNet}

ResNet  \cite{He2016} is a well-established architecture in the deep learning community, known for winning the ImageNet competition (ILSVRC, 2015) for classification tasks. This model solved the problem of vanishing gradients for learning models with many layers of convolutional network. One of the observations that was made with AlexNet networks was that the deeper the network, the better the performance. However, it was found that beyond a certain depth, performance deteriorates. The error gradient through the back propagation tends to cancel it out. To ensure that the information is not lost during the backpropagation, the authors invented a particular architecture that implement residual connections. The structure of the net is organized into several blocks, each one composed of two or three layers of convolutions, associated with a batch normalization layer and a ReLU activation. Thus, the stacking of these blocks allows the construction of deeper convolutional learning models by reducing the effect of the vanishing gradient. Several ResNet architectures have been proposed where each block uses two convolutions for shallower models like ResNet18 or ResNet34, or three convolutions for deeper one like ResNet50, ResNet101and ResNet152. 

\section{Sargassum Level Classification}
\label{sec:sargassum classification}

Our target is to provide an estimation of the sargassum amount that is present in a landscape photography of the beach. Given that the photos are taken from monocular cameras with different specifications, it is not possible to directly estimate volumes from those un-calibrated images. Therefore, we propose to use a supervised learning approach, where several known examples are given to a model and that model is trained to make predictions on unknown examples. In particular, we have modeled the problem as a classification problem, where given an image $I$; a model $\Phi$ assigns it a label $l$ from a set of predefined labels, namely:

\begin{equation}
    \Phi(I): R^{w \times h} \rightarrow L = \{ l_1 \dots l_n \}
\end{equation} where $w$ and $h$ define image's width and height respectively, and $n$ is the quantization of the sargassum level. Below, we describe the dataset that we have built, the proposed sargassum quantization and the implementation of the model by deep convolutional neural networks.

\subsection{Sargassum Dataset}

\begin{figure}[tb]
\centering
 \subfloat[][beach]{
   \includegraphics[width=0.22\textwidth]{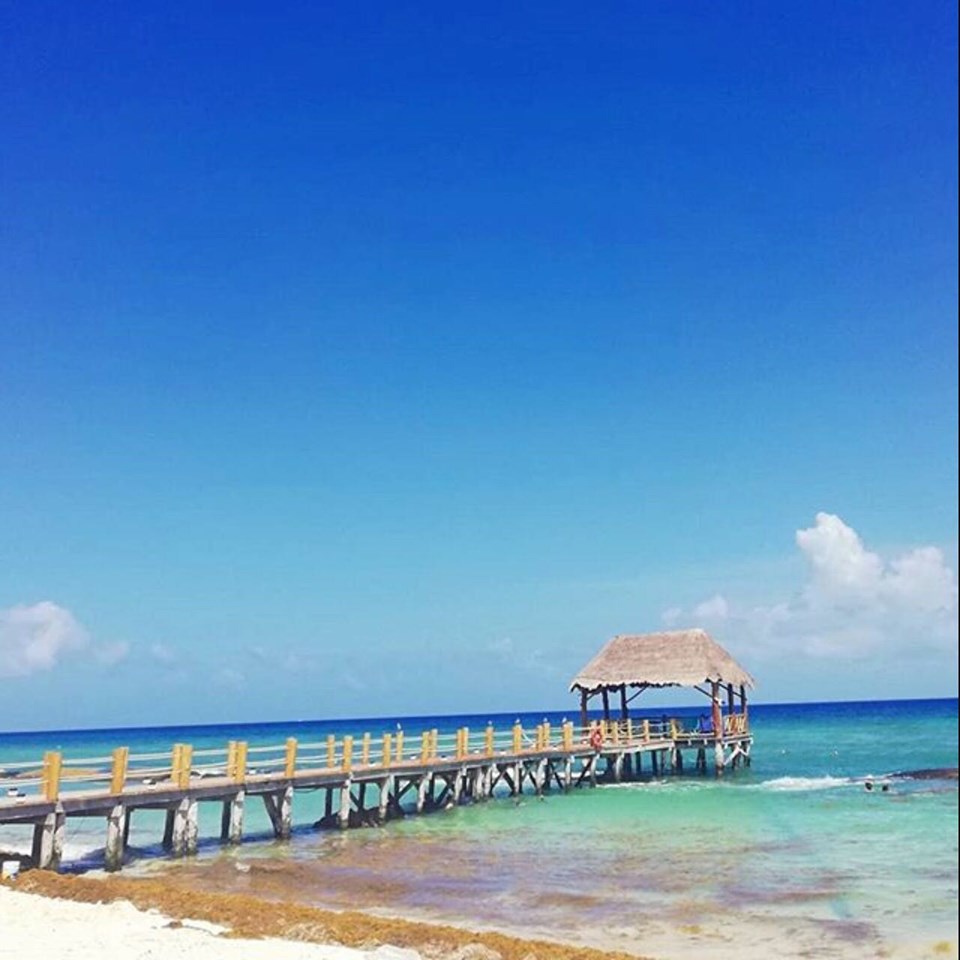}
 }
 \subfloat[][beach]{
   \includegraphics[width=0.22\textwidth]{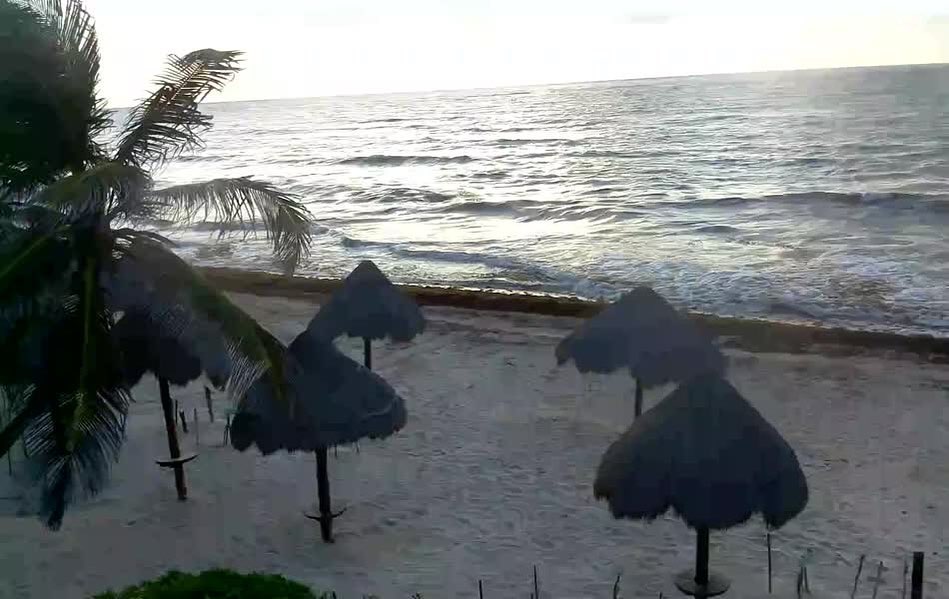}
 }
  \subfloat[][beach]{
   \includegraphics[width=0.22\textwidth]{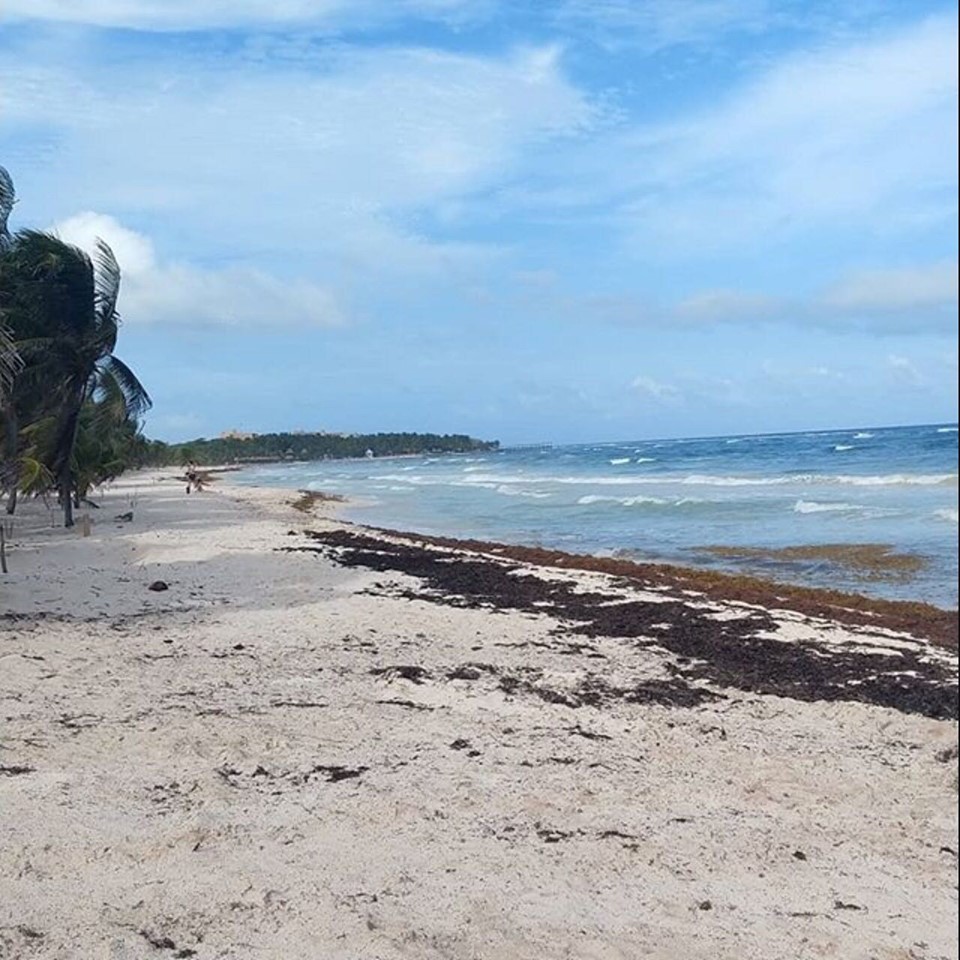}
 }
 \subfloat[][beach]{
   \includegraphics[width=0.22\textwidth]{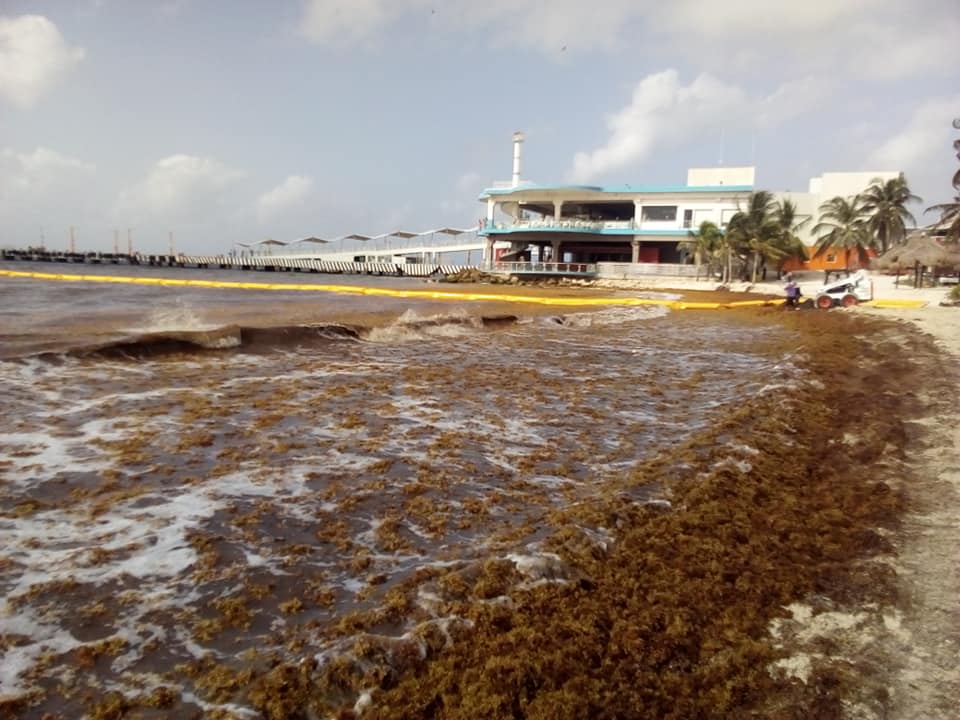}
 }
 
 \subfloat[][sea]{
   \includegraphics[width=0.22\textwidth]{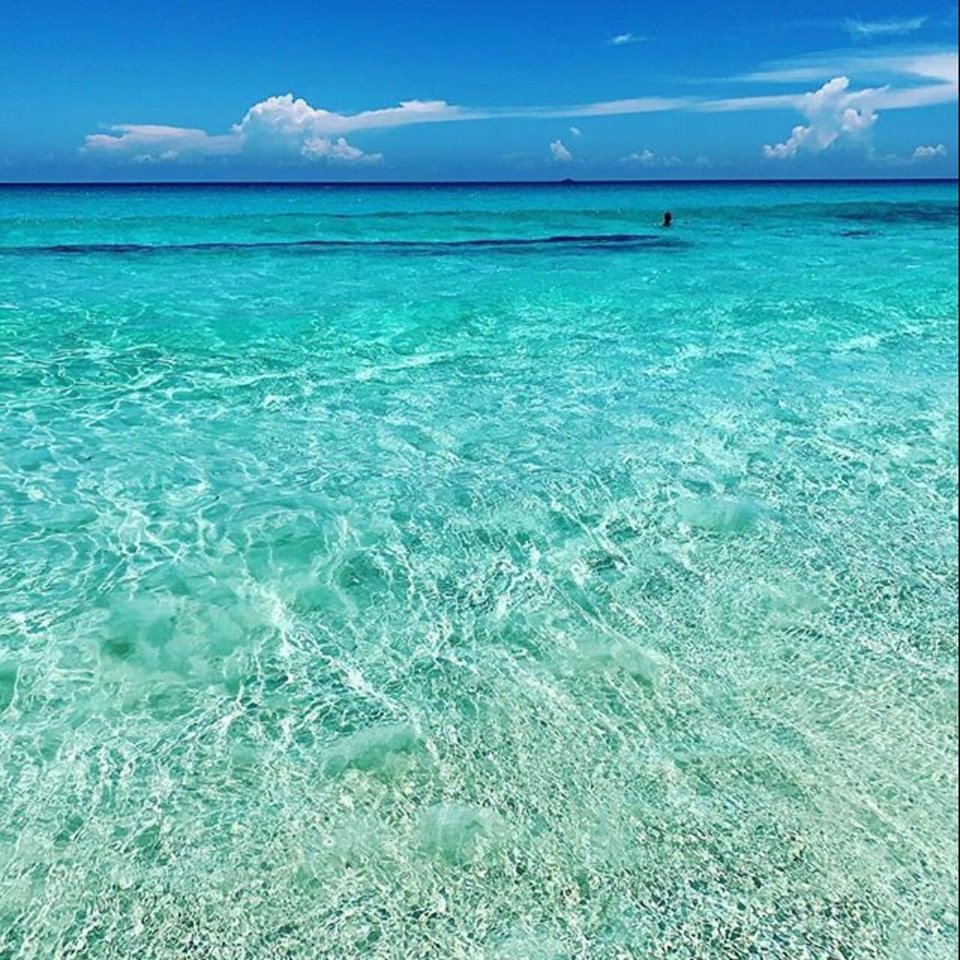}
 }
 \subfloat[][beach]{
   \includegraphics[width=0.22\textwidth]{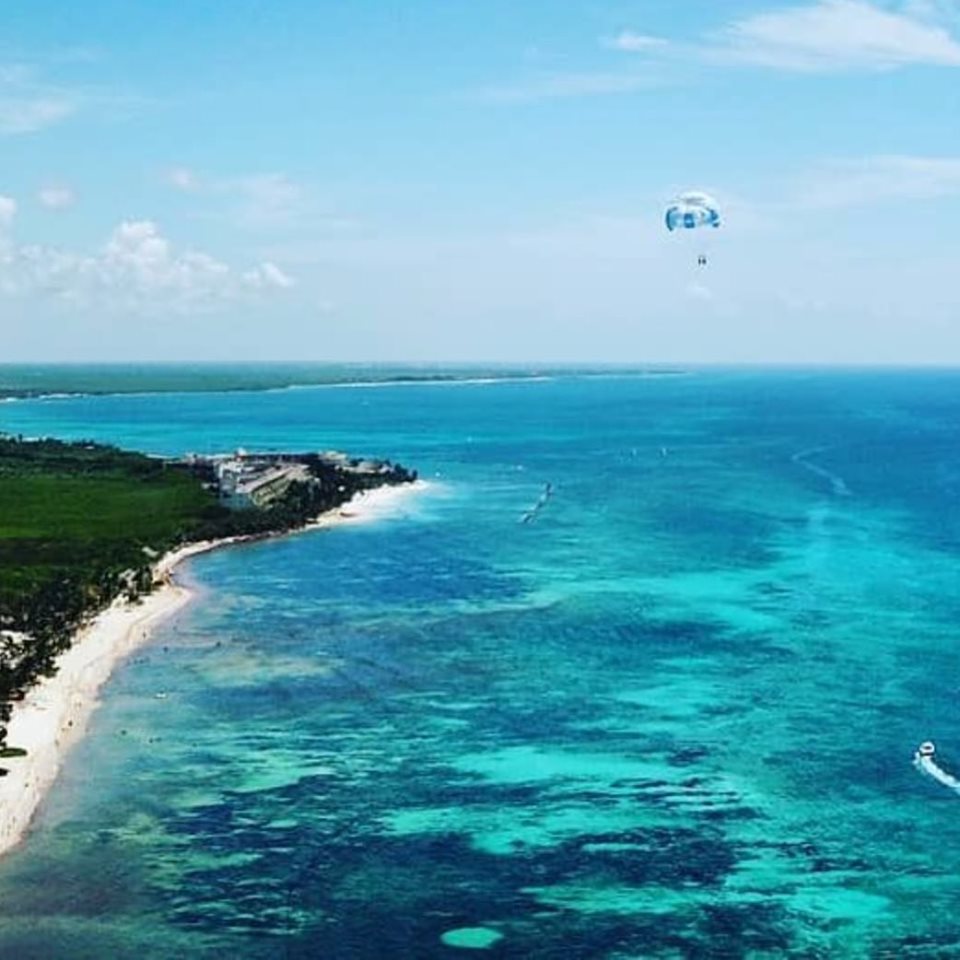}
 }
  \subfloat[][aerial]{
   \includegraphics[width=0.22\textwidth]{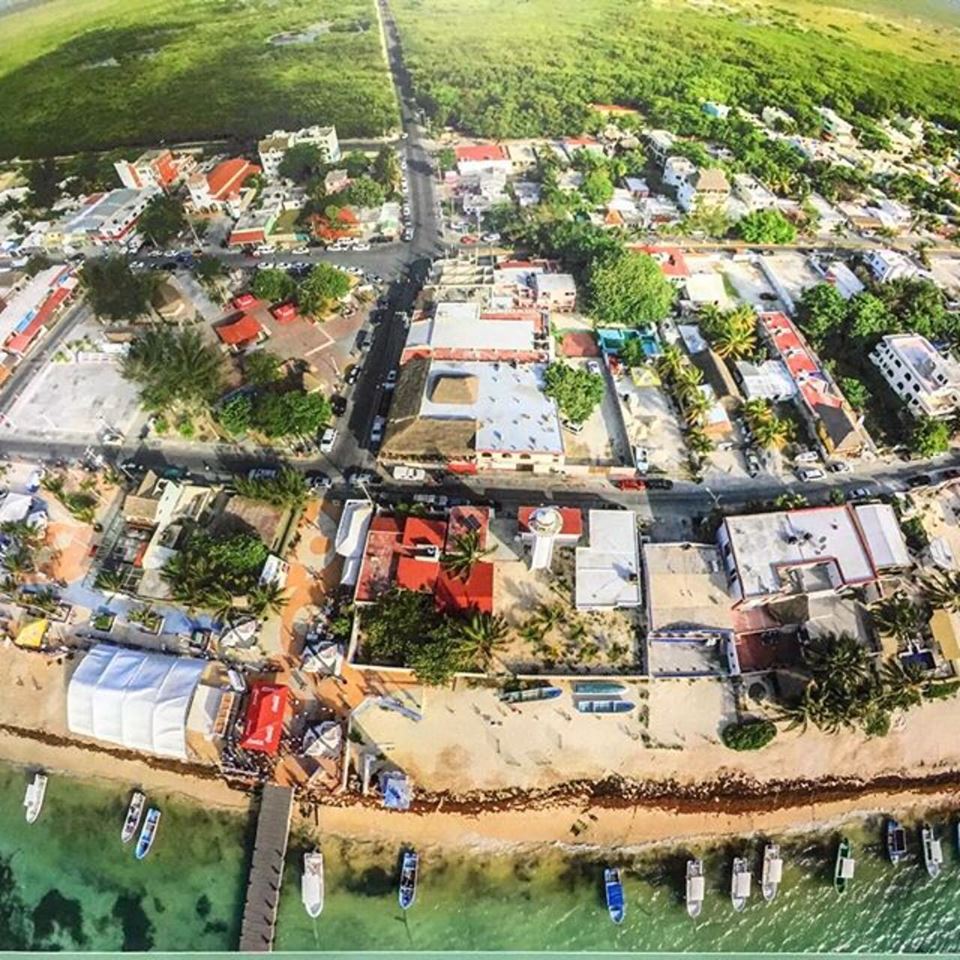}
 }
 \subfloat[][beach]{
   \includegraphics[width=0.22\textwidth]{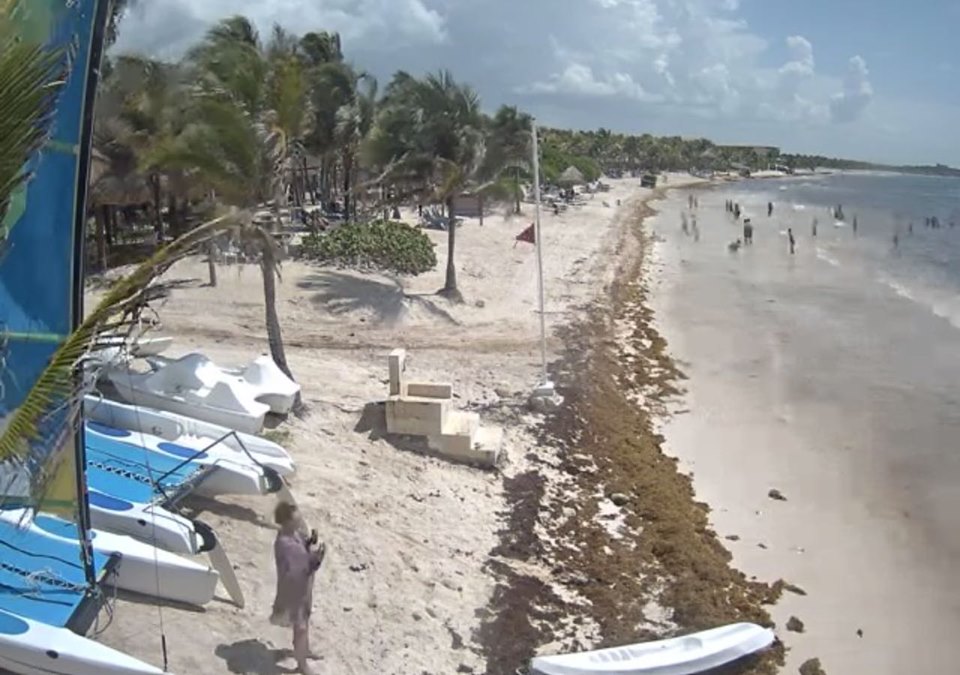}
 }
 \caption{Image examples from the constructed dataset.}
 \label{fig:examples}
\end{figure}

\begin{figure}[tb]
    \centering
    \includegraphics[width=0.6\linewidth]{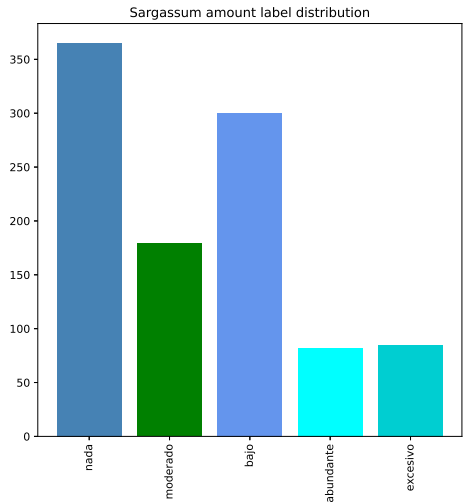}
    \caption{Distribution of labels for characterizing the amount of sargassum in images. The labels were defined using spanish words. They are \textit{nada} (nothing), \textit{bajo} (low), \textit{moderado} (mild), \textit{abundante} (plenty) and \textit{excesivo} (excesive).}
    \label{fig:amount_label_distribution}
\end{figure}

Our dataset was built by gathering images from public publications in Facebook and Instagram social networks. The dataset was first stored locally and manually labeled. If the reader is interested in replicating our results, we provide a public version of the dataset at \cite{vasquezSargazoDS}, however, we only include the features that were extracted by the neural networks and we do not provide the pictures in order to keep people privacy (even though that those pictures are publicly available). Fig. \ref{fig:examples} shows some picture examples from the dataset. The images were stored using the properties from their original posts, namely, resolution and size varies between images. So far, 1011 images have been gathered. Four labels have been attached to each image: place, date, level and scene. 'level' provides a manually assigned level indicating the amount of sargazo found in that image. The levels were stored using spanish words and following an incremental order: $L = \{$ \textit{nada} (nothing) , \textit{bajo} (low), \textit{moderado} (mild), \textit{abundante} (plenty) and \textit{excesivo} (excesive) $\}$. We used spanish words to facilitate the work of the human labelers. With respect to level labels, the dataset is imbalanced given that more examples from `nothing' and `low' levels were gathered. Please see Fig. \ref{fig:amount_label_distribution} where the distribution of labels is displayed. Scene label provides information about from where the image was taken; the labels are: \textit{playa} (beach), \textit{mar} (sea), \textit{tierra} (land), \textit{aérea} (aerial). Place and date labels indicate where and when the images were taken. 

\subsection{Classification with Deep Convolutional Neural Networks}

Recently, deep neural networks have proven to effectively solve several complex computer vision tasks. One advantage of such discoveries is the fact that the knowledge acquired in some context can be moved to a similar domain by using transfer learning. Transfer learning reuse the learned network weights for initializing a custom network that is intended to solve the target task. In this work, the target task is the classification of sagassum level while the source task is the Imagenet classification \cite{ILSVRC15}. 

There are at least two ways for using transfer learning: feature extraction and fine tuning. Feature extraction keeps the feature extraction capabilities of an already trained network by freezing the convolutional layer weights while a new classifier is attached to the feature extraction; then the new classifier is trained according to the target task. On the other hand, fine tuning replaces the classifier according to the target task and then the full network is fitted to the target task. Our hypothesis is that transfer learning can provide a good approximation of the sargassum classification. In the next section, we present several experiments in order to validate our proposal.

\section{Experiments}
\label{sec:experiments}

\subsection{Network architectures}

In this experiment, we test several state-of-the art networks for the classification task using all the levels in the dataset. The tested networks are: Alexnet \cite{Krizhevsky2012}, GoogleNet \cite{Szegedy2015}, VGG16 \cite{SimonyanZ15} and RestNet18 \cite{He2016}. The objective of this experiment is to characterize the effectiveness of the network architectures by validating them under three training paradigms: feature extraction (F.E.), fine tuning (F.T.) and training from 'scratch' (T.S.). 

The parameters used are the following: epochs = 500, learning rate = 0.001 and batch size = 100. The batch size was selected as the largest amount of images that can be loaded into memory during training. The training was carried out online on a Kaggle's virtual machine with GPU and 15 GB of memory RAM. Training time was 3 hours in average. 80\% of the dataset was used for training while 20\% was selected for validation. The best CNN weights corresponds to the epoch were the best validation accuracy was reached.

We can observe, from the results summarized in table \ref{tab:nets}, three facts. First, the problem addressed in this work is quite different from the typical image classification problems, this is observed in the low validation accuracy (below 60\%); our hypothesis is that learned features from Imagenet are not enough for this problem and that the current amount of examples is small for extracting such features. Two, fine tuning has shown the best performance during validation; this phenomenon supports our hypothesis that the features from Imagenet does not define the sargassum because the net is learning to extract new features. Third, VGG16 has shown the best performance; we think that given the amount of available examples, the depth of VGG16 is adequate for learning features and for beating the other networks, while at the same time, it is also shallow for avoiding overfit which can be occurring with ResNet18.

\begin{table}[tb]
    \centering
    \begin{tabular}{|c|c|c|c|c|}
    \hline
        Network & F.E. & F.T. & T.S.   \\
    \hline
        AlexNet & 0.522 & 0.556 & 0.467 \\ 
    \hline
        GoogleNet &0.485 & 0.4752 &0.436 \\
    \hline
        ResNet18 & 0.566 & 0.5860 & 0.512 \\
        \hline
        VGG16 & 0.522 & \textbf{0.5862} & 0.527 \\
        \hline
    \end{tabular}
    \caption{Validation accuracy reached by each network. Testing variants are transfer learning F.E., fine tuning F.T. and training from scratch T.S.}
    \label{tab:nets}
\end{table}




 \begin{figure}[tb]
    \subfloat[Confusion matrix\label{subfig-1:dummy}]{%
      \includegraphics[trim=3cm 0cm 0.5cm 0.9cm, clip=true,width=0.45\textwidth]{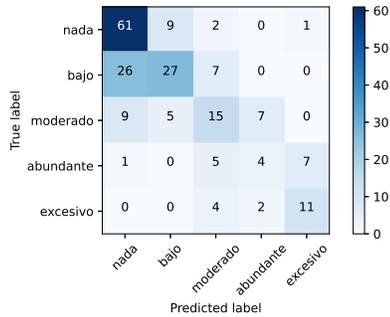}
    }
    \hfill
    \subfloat[Normalized confusion matrix\label{subfig-2:dummy}]{%
      \includegraphics[trim=3cm 0cm 0.5cm 0.9cm, clip=true, width=0.45\textwidth]{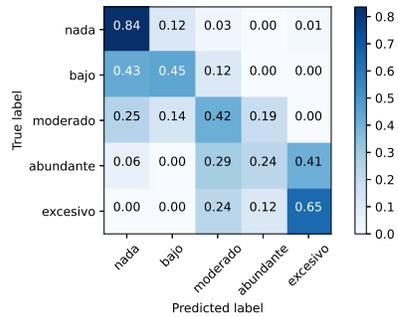}
    }
    \caption{Single and normalized confusion matrices for the classification of sargassum. Translated labels are: \textit{nada} (nothing), \textit{bajo} (low), \textit{moderado} (fair), \textit{abundante} (plenty), \textit{excesivo} (excesive).}
    \label{fig:confusion_matrices}
 \end{figure}
 
  \begin{figure}[tb]
    \subfloat[Nothing/nothing \label{subfig-1:e}]{%
      \includegraphics[trim=3.8cm 1.1cm 3.5cm 1.1cm, clip=true,width=0.25\textwidth]{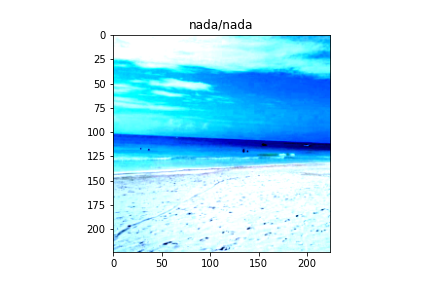}
    }
    \subfloat[Nothing/nothing \label{subfig-2:e}]{%
      \includegraphics[trim=3.8cm 1.1cm 3.5cm 1.1cm, clip=true, width=0.25\textwidth]{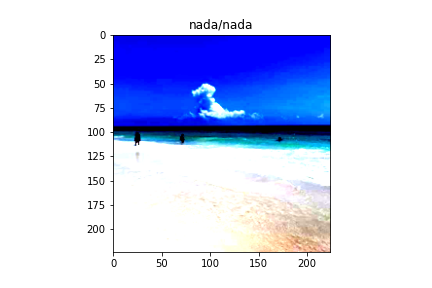}
    }
    \subfloat[Nothing/nothing\label{subfig-3:e}]{%
      \includegraphics[trim=3.8cm 1.1cm 3.5cm 1.1cm, clip=true,width=0.25\textwidth]{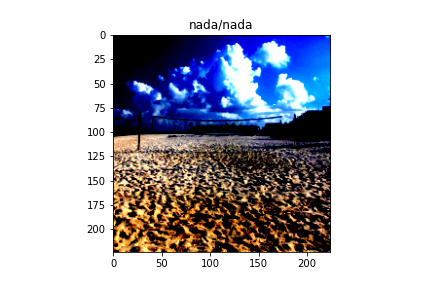}
    }
    \subfloat[Excesive/excesive\label{subfig-4:e}]{%
      \includegraphics[trim=3.8cm 1.1cm 3.5cm 1.1cm, clip=true, width=0.25\textwidth]{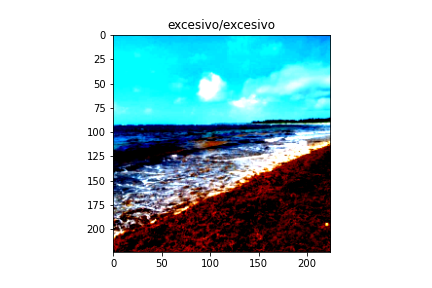}
    }
    
    \subfloat[Low/nothing\label{subfig-5:e}]{%
      \includegraphics[trim=3.8cm 1.1cm 3.5cm 1.1cm, clip=true,width=0.25\textwidth]{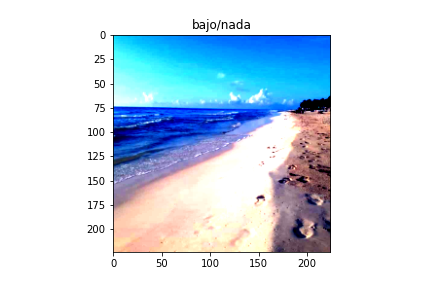}
    }
    \subfloat[Plenty/plenty\label{subfig-6:e}]{%
      \includegraphics[trim=3.8cm 1.1cm 3.5cm 1.1cm, clip=true,width=0.25\textwidth]{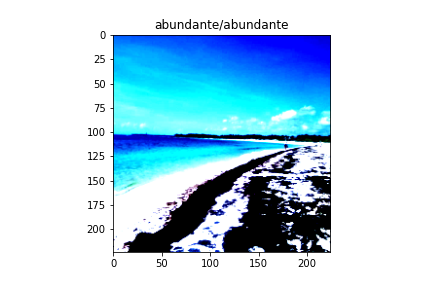}
    }
        \subfloat[Low/mild\label{subfig-7:e}]{%
      \includegraphics[trim=3.8cm 1.1cm 3.5cm 1.1cm, clip=true,width=0.25\textwidth]{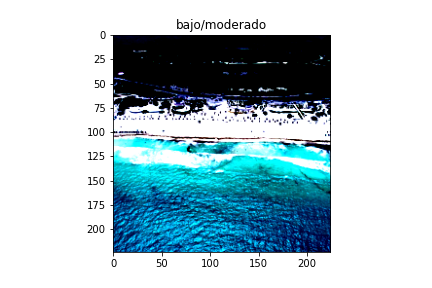}
    }
        \subfloat[nothing/fair\label{subfig-8:e}]{%
      \includegraphics[trim=3.8cm 1.1cm 3.5cm 1.1cm, clip=true,width=0.25\textwidth]{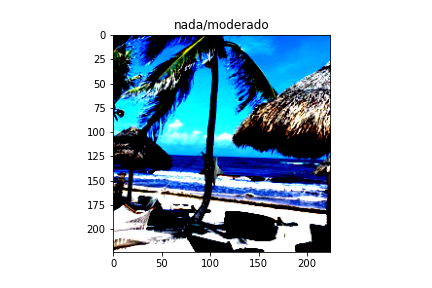}
    }
    \caption{Examples of sargassum level classification using the selected convolutional neural network. Displayed images are already pre-processed: normalization and resize. The label for each example indicates: 'predicted label / true label'.}
    \label{fig:pred_examples}
  \end{figure}

\subsection{Predictions analysis}

In this experiment, we make a deeper analysis on the predictions provided by the model with the highest validation accuracy observed (VGG16 trained with fine tuning). From Fig. \ref{fig:confusion_matrices}, we can observe that predictions are near to the true label. This phenomenon is better observed in the normalized matrix (Fig. \ref{subfig-2:dummy}), where the majority of predictions are in the diagonal (true positives) and the remaining ones are distributed among the contiguous classes. This reflects the fact that there is no a hard rule for this classification, but it is based on manual labeling which is prone to the point of view of the human labeler. See Fig. \ref{fig:pred_examples} for prediction examples. In conclusion, the predictions provided by tested models can be useful as a good approximation of the sargassum amount. 


\section{Conclusions and Future Work}
\label{sec:conclusions}

A study for sargassum level classification in public photographs has been presented. Several networks have been tested using different training strategies. The winning network has been VGG16 trained under fine-tuning. The results show an accuracy of 58 percent. We hypothesize that this limited accuracy is because the levels were assigned using human criteria. In consequence, a discrepancy between labels can be found. A hard rule for assigning labels could increase the reached validation accuracy. On the other hand, the prediction distribution is narrow, therefore, predicted labels make sense to human observers. We hope that our system can offer a tool for sargassum monitoring. So that, scientists, ecologists and authorities can have a better understanding of the phenomenon. As a result, cleaning efforts and studies can be directed to critical areas. Our next research direction is to segment the sargassum from the images and to include image geo-localization.

\subsection*{Acknowledgements}

J.I. Vasquez thank to CONACYT for the \textit{Cátedra} 1507 project. The authors thank to the following people who have recolected images from public forums: Miguel Sanchez, Alberto Flores and Cesar Pérez. 

\subsection*{Data availability}

All gathered data is open access and available via Kaggle:  \url{https://www.kaggle.com/irvingvasquez/sargazo-dataset}. The dataset will be continuously updated as new data is obtained.

\subsection*{Author contributions}
JIV conceived the ideas and designed methodology; AGF and EV analysed the data; HT and AU led the writing of the manuscript. All authors contributed critically to the drafts and gave final approval for publication.



\bibliographystyle{apalike} 
\bibliography{cas-refs}





\end{document}